\documentclass[journal]{IEEEtran}
\ifCLASSINFOpdf
\else
\fi

\usepackage{times}
\usepackage{epsfig}
\usepackage{graphicx}
\usepackage{amsmath}
\usepackage{amssymb}
\usepackage{caption2}
\usepackage{balance}
\usepackage{multirow}
\usepackage{bm}
\usepackage{mathrsfs}
\usepackage{setspace}
\usepackage{cite}

\usepackage{algorithm}
\usepackage{algpseudocode}
\usepackage{amsmath}

\begin{document}
%
\title{Bilateral Cross-Modality Graph Matching Attention for Feature Fusion in Visual Question Answering}
%
%
%

\author{JianJian~Cao, Xiameng Qin,
        Sanyuan Zhao, and~Jianbing Shen,~\IEEEmembership{Senior Member,~IEEE}


\IEEEcompsocitemizethanks{
\IEEEcompsocthanksitem JianJian~Cao and Sanyuan Zhao are with Department of Computer Science, Beijing Institution of Technology, Beijing, China.
Email: caojianjianbit@gmail.com, zhaosanyuan@bit.edu.cn. 
\IEEEcompsocthanksitem Xiameng Qin is with Baidu Inc., Beijing, China. Email: qinxiameng@baidu.com.
\IEEEcompsocthanksitem Jianbing Shen is with the Inception Institute of Artificial Intelligence, Abu Dhabi, UAE.
(Email: shenjianbingcg@gmail.com)
\IEEEcompsocthanksitem Corresponding author: \textit{Sanyuan Zhao}
\IEEEcompsocthanksitem This work was supported by the National Natural Science Foundation of China under Grant 61902027.
}
}

%
%

\markboth{IEEE Transactions on Neural Networks and Learning Systems}%
{Jian-Jian \MakeLowercase{\textit{et al.}}: Bilateral Cross-Modality Graph Matching Attention for Feature Fusion in Visual Question Answering}
%



\maketitle

\begin{abstract}
Answering semantically-complicated questions according to an image is challenging in Visual Question Answering (VQA) task.
Although the image can be well represented by deep learning, the question is always simply embedded and cannot well indicate its meaning.
Besides, the visual and textual features have a gap for different modalities, it is difficult to align and utilize the cross-modality information.
In this paper, we focus on these two problems and propose a Graph Matching Attention (\textit{GMA}) network.
Firstly, it not only builds graph for the image, but also constructs graph for the question in terms of both syntactic and embedding information.
Next, we explore the intra-modality relationships by a dual-stage graph encoder and then present a bilateral cross-modality graph matching attention to infer the relationships between the image and the question.
The updated cross-modality features are then sent into the answer prediction module for final answer prediction.
Experiments demonstrate that our network achieves state-of-the-art performance on the GQA dataset and the VQA 2.0 dataset. The ablation studies verify the effectiveness of each modules in our \textit{GMA} network.
\end{abstract}


%
\IEEEpeerreviewmaketitle

\section{Introduction}
    \IEEEPARstart{G}{iven} a visual image, the goal of Visual Question Answering (VQA) is to correctly answer a relative natural language question. It requires a comprehensive cross-modality understanding and visual-semantic reasoning between images and questions. VQA plays an important role and has a wide potential application in the future life, such as virtual assistants and autonomous agents.
    Therefore, it has aroused extensive interest from both vision and natural language communities.
    With the explosive progress of deep neural networks (DNN), a large amount of deep learning methods, \emph{i.e.}  BUTD~\cite{Anderson_2018_CVPR} and Pythia~\cite{DBLP:journals/corr/abs-1807-09956}, have been developed for better representation learning. These feature representations have been introduced into VQA and efficiently improved the performance of existing algorithms.
    \begin{figure}[htp]
        \begin{center}
        \centering
        \includegraphics[width=1.0\linewidth]{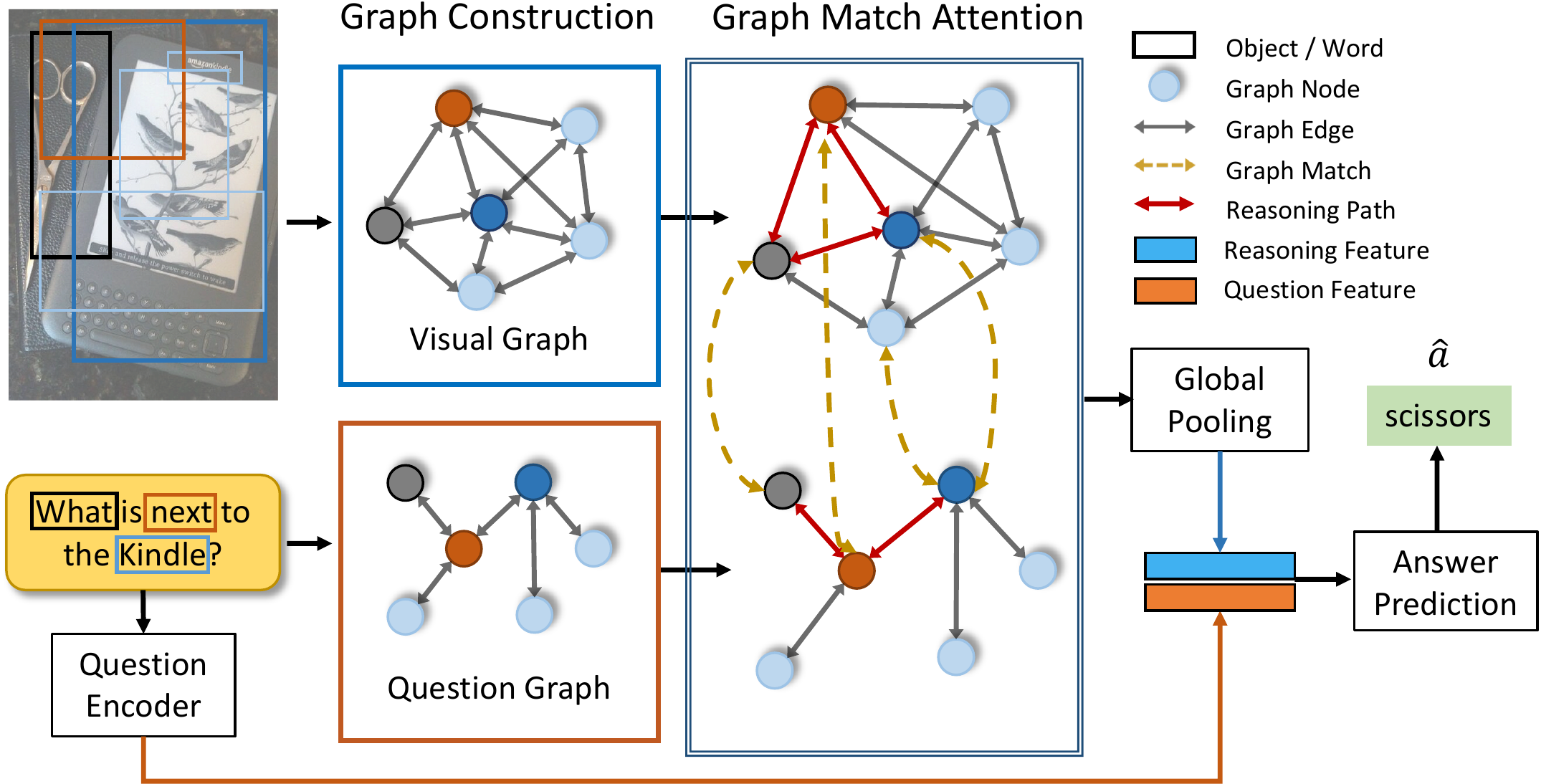}
        \end{center}
      \caption{The pipeline of our \textit{GMA} network for the VQA task.
      Both the visual graph and the question graph are constructed as the input of \textit{GMA} network. Consequently the intra-modality features are explored and the cross-modality information are aligned. The reasoning feature, output from the \textit{GMA} network, is merged with the question feature to predict final answer.
      }
        \label{fig1}
    \end{figure}
    In recent years, plenty of VQA studies focused on relational reasoning. The attention mechanism~\cite{DBLP:journals/corr/VaswaniSPUJGKP17}, which is capable to encapsulate the most relevant information from the image or the question, has been adopted for relation reasoning.
    BAN~\cite{NIPS2018_7429} and DFAF~\cite{Gao_2019_CVPR} designed a variety of attentions to explore the implicit relationship between the image and question domain, and their effects are remarkable. These VQA frameworks adopted simple attention mechanisms, learning the "black-box" relationships in an agnostic dimensional feature space without underlying restrictions. The arbitrariness increased the risk of over-fitting and limited high-level relational reasoning. Besides, some recently works~\cite{DBLP:journals/corr/abs-1812-09681,Wang_2019_CVPR} concentrated on learning an explicit visual semantic relationship in an image via Visual Relation Detection (VRD)~\cite{DBLP:journals/corr/abs-1902-10200,Yu_2017_ICCV}. To some extent, this kind of frameworks put forward a new possible direction for solving the relational reasoning in the VQA task, but they increased uncertainty of the model, making the performance heavily dependent on the relationship learning algorithms.
    In addition, predominant approaches of VQA focused on learning a joint cross-modality representation to represent the alignment between the image region and the corresponding question.
    Specifically, bilinear fusion~\cite{DBLP:journals/corr/GaoBZD15} exploited bilinear pooling to capture pairwise interactions among the modalities.
    Some sophisticated schemes, including MCB~\cite{DBLP:journals/corr/FukuiPYRDR16}, MLB~\cite{KimHadamard} and MUTAN~\cite{Ben-younes_2017_ICCV}, which designed different linear fusion strategies, demonstrated better performances and less computational burden than Bilinear Fusion.
    Nevertheless, modern fusion methods carried out feature alignment in the absence of modal structure (e.g. positional relationships of the image regions, and grammatical relationships of the question) and output one monolithic vector unable to describe fine-grained information for both visual and language semantics.
    As a result, the alternatives almost produced a sub-optimal solution leading to imperfect cross-modality feature representation.

    The graph method is generally known as using a topological data structure to express complex relationships between nodes. In VQA, several graph-based approaches~\cite{DBLP:journals/corr/abs-1806-07243,YangScene} were proposed and shown great improvement in visual understanding~\cite{2020_Graph}. They leveraged topology to the subtle context information in order to remedy the defect of Euclidean distance in relationship inference. The drawbacks lie in the ignorance of the syntactic relationship of the language modality and mere consideration of visual information. This is a serious problem for VQA model understanding.

    \begin{figure*}[htp]
    \begin{center}
    \centering
    \includegraphics[width=0.95\textwidth]{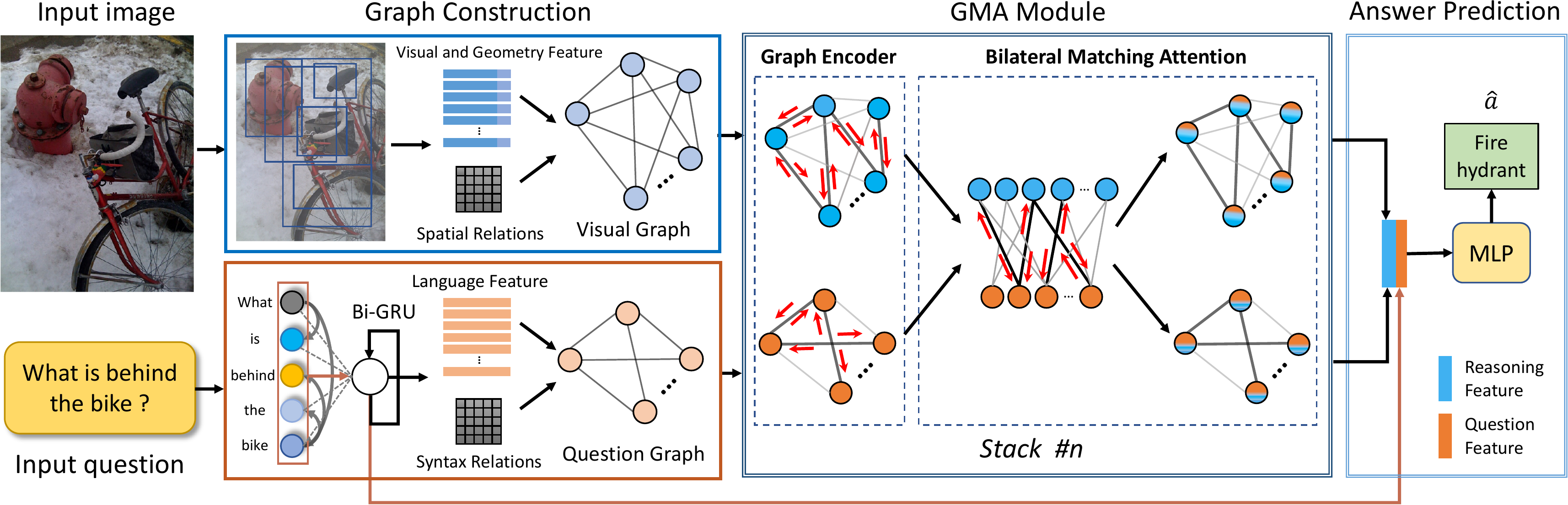}
    \end{center}
    \caption{The architecture of our Graph Matching Attention network.
    The visual graph is constructed by the apparent and geometry features of the visual objects and their spatial relations.
    The question graph is built with language features and the syntax tree extracted from the question.
    According to the two graphs, the~\textit{GMA} module aligns and fuses the cross-modality features by the graph encoder and the bilateral matching attention sub-modules. The~\textit{GMA} features are further fed into the answer prediction module to carry out answer prediction.
    }
    \label{fig2}
    \end{figure*}


    In this paper, we propose a Graph Matching Attention (\textit{GMA}) network to solve the problems of unsuitable question description and the alignment between the visual and textual modality information.
    As shown in Fig.\ref{fig1},
    our network constructs two graph data structures, to depict the positional geometry of image regions and syntactic structure of the question, respectively.
    In the view of the topological information, the graph patterns for both the two modalities bring about convincing relational reasoning in cross-modality understanding.
    To solve the shortcoming of the traditional attention mechanisms applied in VQA, we propose the \textit{GMA} module to develop a more intuitive relational reasoning. Based on the visual and textual graphs, the \textit{GMA} module guides to align the attention across the two modalities.
    We firstly explore information relationships inside the visual and textual modalities respectively, by a dual-stage graph encoder to represent the relationships of the nodes explicitly and implicitly. The graph encoder in our \textit{GMA} produces meaningful internal relationships in the image and the question, separately, improving the attention interpretation.
    Then, the intra-modality features are forward into the graph matching attention modules to align and pass information between visual and textual features in a bilateral manner.
    By practicing the graph matching between the visual and texture graph patterns, each node is capable to adopt a feature fusion inference from the multi-modality representations, rather than a monolithic feature in previous works.
    Furthermore, the \textit{GMA} contributes to the performance with the participation of multi-modality structural information.
    Finally, each graph feature is weighted by the bilateral cross-modality attention and sent to the answer prediction module to calculate the score for the potential answers.

    Specifically, as shown in Fig.\ref{fig1}, the \textit{GMA} utilizes the Graph Convolutional Networks (\textit{GCNs}) to successively capture the explicit and implicit relations among the nodes inside each modal (the red arrows in Graph Match Attention module).
    These relational features reveal fine-grained intra-modality concepts and provide a holistic view of modal interpretation.
    Next, a pairwise matching (the yellow dashed arrows linking visual and question graph nodes) is applied to inter-modality nodes, and attaches the semantic alignment.
    The graph matching operations dynamically update the nodes and pass information flows within each modality to capture higher-order inter-modality relations. In this way, every node aggregates the attention-weighted information from the other instances under the guidance of graph structure. Besides, the \textit{GMA} modules can be stacked and explore a deeper level semantic relationship.

    The main contributions of our work are four-fold:
    \begin{itemize}
    \item We propose to construct graphs for both the image regions and the questions, respectively, to explore their structured intra-modality information and unify their representation for further alignment.

    \item We design a novel Graph Matching Attention (\textit{GMA}) module which contains a  dual-stage graph encoder and a bilateral cross-modality graph  matching attention. Inheriting graph learning, this module introduces the graph matching operation to solve the cross-modality alignment and feature fusion problem, and learns inter- and intra-modality relations.

    \item By cascading several \textit{GMA} modules, the network can learn the deeper relations and achieve high-level relational reasoning, which makes the VQA inference more reasonable.
   	
   \item We conduct ablation studies and compare our network with state-of-the-art methods on the famous GQA and VQA 2.0 datasets. The results convince the effectiveness of our proposed \textit{GMA} module.
   \end{itemize}

\section{Related Work}
    \subsection{Visual Question Answering}
    Visual Question Answering (VQA) is a challenging problem, which involves two fields of Computer Vision and Natural Language Processing (NLP). The VQA task~\cite{Antol_2015_ICCV} requires that the trained model can answer any natural language questions of a given image.
    In real application scenarios, it usually becomes a multimedia answering task~\cite{2011Multimedia}, and can also be directly applied to the multi-modal dialog system~\cite{DBLP:conf/mm/HeLLCXHY20}.
    Current algorithms~\cite{Yang2016Stacked,LuHierarchical,Anderson_2018_CVPR,NIPS2018_7429} for VQA task are usually designed as a multi-label classification problem, which makes the neural network infer answers in a limited set of classes. The current mainstream VQA framework is usually composed of image encoder, question encoder, multi-modal features fusion module, and answer predictor. Feature learning is an essential part of VQA algorithms. Due to the success of deep representation learning~\cite{Simonyan2014Very, He2016Deep}, BUTD~\cite{Anderson_2018_CVPR} derived from Faster RCNN~\cite{NIPS2015_5638}, the performance of VQA has been greatly improved.
    Recently, the researchers pay more attention to the relational reasoning of VQA task. Specifically, Murel~\cite{Cadene_2019_CVPR} proposed to stack several layers of attention to gradually focus on the most important regions.
    ReGAT~\cite{ReGAT_LI_2019} proposed graph-based attention mechanisms to learn the relational semantics between the visual and language modalities.
    Stacked attention~\cite{Yang2016Stacked} and Co-attention~\cite{LuHierarchical} based approaches inferred visual and textual attention distributions for each modality. BAN~\cite{NIPS2018_7429} inferred a bilinear attention distribution between two modalities. DFAF~\cite{Gao_2019_CVPR} integrated cross-modality self-attention and co-attention mechanisms to achieve effective information flows within and between two modalities.
    Besides, some recent algorithms attempted to solve the problem of multi-modality feature fusion in VQA. MCB~\cite{DBLP:journals/corr/FukuiPYRDR16}, MLB~\cite{KimHadamard} and MUTAN~\cite{Ben-younes_2017_ICCV} designed different bilinear fusion strategies to fuse the cross-modality features. In addition, some popular methods\cite{inbook, 2020Counterfactual} begin to design different data augmentation algorithms to enhance the effect of VQA model.
    Those methods have proved useful for the VQA task, but there are still some unsolved issues, such as the alignment between the two modalities and the absence of the structural information of each modality.
    In this paper, we construct two graphs to respectively represent the geometry relationship of image regions and the syntactic structure of its question. A graph matching attention module is designed to successively align and fuse the visual and textual features, in order to learn the semantic representation for final answer prediction.

    \subsection{Relational Reasoning}
    Recently, the deep learning community began to deal with complex relational reasoning problems, such as relationship detection~\cite{YIN_2021_TPAMI}, object recognition~\cite{DBLP:journals/corr/abs-1901-03067}, and abstract reasoning~\cite{Moore1981Abstract}.
    ~\cite{2019Hierarchical} proposed the HDWE model to address click feature prediction from visual features for Fine-grained Image Recognition. ~\cite{2019Spatial} proposed SPE-VLAD layer to reflect the structural information of the images for place recognition.
    ~\cite{Andreas2016Neural} proposed  multi-step reasoning models NMNs that built question-specific layouts. Based on NMNs, Stack-NMN ~\cite{Hu2018Explainable} algorithm was proposed to practice more effective multi-step reasoning. FiLM~\cite{Ethan_2017_FiLM} modulated an image representation with the given question via conditional batch normalization, and was extended by~\cite{Yao_2018_Cascaded} with a multi-step reasoning procedure, where both modalities can modulate each other. MAC~\cite{Drew_2018_MAC} performed multi-step reasoning while recording information in its memory. QGHC~\cite{Gao_2018_ECCV} predicted question-dependent convolution kernels to modulate visual features. In addition, recently proposed methods~\cite{NIPS2018_7429,Gao_2019_CVPR} applied various attention mechanisms in reasoning framework to enhance the learning of implicit relational semantics. Given a question, these models assigned an importance score to each region, and employed them to weight-sum pool the visual representations. The attention mechanism has been proved to be effective in solving the problem of relational reasoning in the VQA task, but only learning the implicit relations is insufficient.
    It is necessary for the VQA model  to learn explicit relations according to the structural information of each modality. In order to enhance the relational reasoning ability, we propose a novel graph match attention mechanism to learn the relational semantics between the image and the question by combining with the designed graph matching method. 

    \subsection{Deep Graph Matching}
    Graph Matching ~\cite{Foggia2014Graph} is a fundamental problem that aims to establish the node correspondences among multiple graphs and has been widely studied in computer vision and pattern recognition areas. To find the maximal similarity between the matched graphs, both the unitary similarity between nodes and pairwise similarity edges from separate graphs are integrated in graph matching methods~\cite{Han_2017_Factorized,NIPS2019_8595}. Naturally, graph matching can encode the high-order geometric information in the matching process so that it can be used to model complex relationships and more diverse transformations.
    Because of the expressiveness and robustness, graph matching relates to many computer vision problems, such as visual tracking, motion recognition, image classification, and so on.
    Recently, researchers began to use the deep features in geometric and semantic visual matching tasks. GMNs~\cite{Li2019Graph} proposed Graph Matching Networks for learning the similarity of graph-structured objects. CLKGM~\cite{Xu_2019_CLKGA} proposed a topic entity graph to represent the knowledge graph context information of an entity and estimated the similarity of two graphs by using the designed attentive-matching method. DLGM~\cite{Zanfir_2018_CVPR} proposed an end-to-end graph matching model that integrates node feature extraction, node/edge affinities learning and matching optimization together in a unified network.
    PCAGM~\cite{Wang_2019_PCAGM} explored Graph Convolutional Networks for graph matching task.
    In this paper, we involve the graph matching algorithms to combine the visual features of the image with the semantic features of the question so as to align the two different feature sub-spaces and learn the high-level cross-modality relations to reasoning the answers of the questions.

\section{Our method}
    As illustrated in Fig.\ref{fig2}, the proposed Graph Matching Attention network consists of a graph construction module, stacked Graph Matching Attention \textit{GMA} modules and an answer prediction module.
    Firstly, we build graph structures for both the image and the question to maximize retention of intra-modality relationships in graph construction module (section \ref{Sec: Graph construction}).
    Then, the \textit{GMA} module (section \ref{Sec: Graph Matching Attention}) explores the cross-modality relationships by carrying bilateral graph matching attention between the visual and the textual graph.
    Finally, the reasoning features from the \textit{GMA} module are aggregated with the textual feature, and fed into the answer prediction module (section \ref{Sec: answer prediction layer}) for the final answer.
\subsection{Graph Construction}
\label{Sec: Graph construction}

\subsubsection{Visual Graph Construction}
    In order to fully express the explicit semantic relationship between objects in visual modality, we construct the visual graph for each image.
    Given an image $\bm{I}$, we build its visual graph $ \mathcal{G}_1 $ according to the object regions and their relationships,
    \begin{equation}
    \begin{aligned}
    & \mathcal{G}_{1} = \{\bm{V}^m, \bm{E}^m\}, \\
    & \left\{\bm{V}^m=\{\bm{v}^m_i\}^{K1}, \bm{E}^m=\{e^m_{ij}\}^{{K_1}\times{K_1}}\right\},
    \end{aligned}
    \end{equation}
    where $\bm{V}^m$ represents the set of graph nodes, $\bm{E}^m$ is the graph edges, $\bm{v}^m_i$ corresponds to a visual object in $\bm{I}$, and $K_1$ is the number of visual graph nodes.
    The visual feature $\bm{V}^m$ can be collected by concatenating the RoI-pooling feature $\bm{R}= \{\bm{r}_i\}, i\in[1,...,K_1]$ from the Faster R-CNN detector~\cite{NIPS2015_5638} and the normalized bounding box coordinates $\bm{B}=\{\bm{b}_i\}, i\in[1,...,K_1]$.
    In order to build the edge connections, dominant literature leveraged a fully-connected graph to learn arbitrary implicit relations under complex attention mechanisms. However, fully-connected architecture loses explicit topology information and inclines to rapidly trapped in over-smoothing during deep graph network training. Therefore,
    The graph edges $\bm{E}^m $ are developed as binary connections, where $e^m_{ij}\in \{0, 1\}$ denotes whether the IoU of the object $i$ and $j$ exceeds a threshold.
    Since the visual objects in an image are independent and $\mathcal{G}_1$ relies on their spatial relations, the visual topology is sparse.

\subsubsection{Question Graph Construction}
    Most of the existing VQA algorithms merely adopted question embedding as the semantic description of the questions and did not incorporate the syntactic patterns, which limited the ability of the VQA model to understand complex questions.
    Motivated by this, we propose to enhance the question representation by both the syntactic structure and internal semantic relevance by a question graph.
    We employ the dependency parsing algorithm~\cite{manning-etal-2014-stanford} to analyze the grammar of question $Q$ and output the syntax dependency tree.
    To be specific, the grammar of question $Q$ is analyzed by dependency parsing and depicted by a syntax dependency tree.
    This tree is regarded as the question graph,
	\begin{equation}
	\begin{aligned}
	& \mathcal{G}_{2} = \{\bm{V}^n, \bm{E}^n\}, \\
	& \left\{\bm{V}^n=\{\bm{v}^n_i\}^{K2}, \bm{E}^n=\{e^n_{ij}\}^{{K_2}\times{K_2}}\right\},
	\end{aligned}
	\end{equation}
    where $K_2$ is the number of words in $Q$, each graph node $\bm{v}^n_i \in \bm{V}^n$ indicates a word and each edge $\bm{e}^n_{ij} \in \bm{E}^n$ represents the syntax relations.
    $\mathcal{G}_2$ is a directed graph and its edges are also binary connections.
    In addition, $\bm{v}^n_i$ is associated with vector embedding, for example, produced by question $Q$ going through a pre-trained GloVe~\cite{pennington2014glove} and a following Bidirectional dynamic Gated Recurrent Unit (Bi-GRU).
    The embedding operation maps the word to a higher dimensional feature space for better contextual representation.
    The question graph fundamentally supports an effective semantic aggregation in the following Graph Matching Attention module which depicted in Sec.~\ref{Sec: Graph Matching Attention}.
    \begin{figure*}[htp]
        \begin{center}
        \centering
        \includegraphics[scale=0.48]{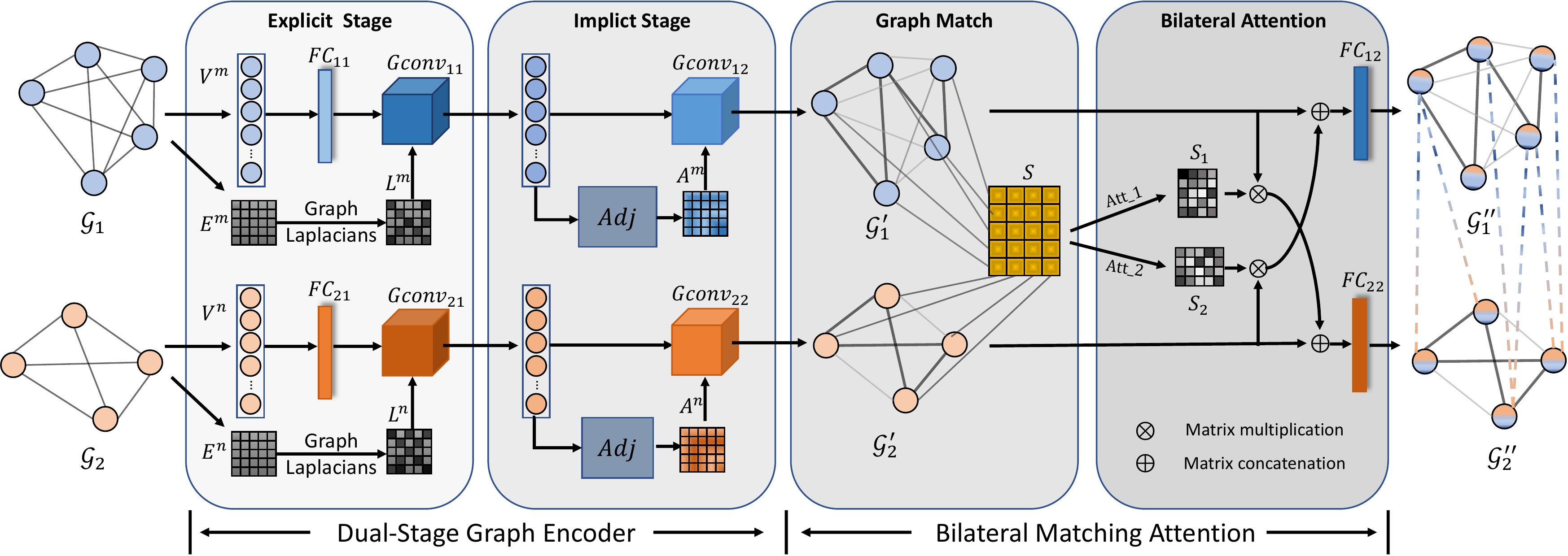}
        \end{center}
        \caption{Illustration of Graph Matching Attention (\textit{GMA}) module.
        Dual-stage graph encoder module  encodes and infers the potential explicit and implicit intra-relations of each graph with graph convolution networks. Bilateral matching attention mechanism are designed to learn inter-relations between the two graph nodes and fuse the cross-modality features.
        }
        \label{fig3}
    \end{figure*}
    \subsection{Graph Matching Attention}
    \label{Sec: Graph Matching Attention}
    Rather than fusing the visual and question features directly in previous works, the proposed Graph Matching Attention (\textit{GMA}) learns the bilateral matching relationships between the visual and textual graphs, and aligns them by passing the cross-modal information to each other.
    Inside the \textit{GMA} module, each graph node can be updated after receiving the cross-modal attention.
    The \textit{GMA} module can be stacked for deeper level of semantic relationship features. Thus, it is convenient for \textit{GMA} model to carry out reasonable cross-modality semantic inference.
    As shown in Fig.\ref{fig3}, the \textit{GMA} module consists of two sub-modules, \emph{i.e.} the Dual-Stage Graph Encoder and the Bilateral Matching Attention module. The details are described as follows.

\subsubsection{Dual-Stage Graph Encoder}
    The inputs of \textit{GMA} module are the node features and the edge matrices of both visual graph and question graph, which have been built in the previous section ~\ref{Sec: Graph construction}.
    The visual graph and question graph are constructed from two different modal spaces and have different topological structures, so there is a gap between them.
    Our dual-stage graph encoder transforms both graph features into the same graph matching subspace.
    Inside the encoder, we apply cascaded graph convolutional layers to successively infer the explicit and implicit relations inside each graph.
    The meaning of explicit stage is information aggregation and relational reasoning between graph nodes based on the clear spatial relationship in visual graph and syntactic relationship in question graph. However, the meaning of implicit stage is to infer according to the similarity relationship between graph node features, which further makes up for the insufficient reasoning of explicit stage.
    As illustrated in Fig.\ref{fig3}, the graph encoder contains two parallel branches to encode visual graph and question graph separately. Each branch is composed of one fully connected ($ FC $) layer and two graph convolutional layers.

    In the explicit relationship inference stage, we firstly utilize the fully connected layer $FC_{11}$ and $FC_{21}$ to transform the features of two graph nodes $ \bm{V}^m $ and $ \bm{V}^n $ into feature matrix $ \bm{X} \in  \mathbb{R}^{K_1 \times d} $ and $ \bm{Y} \in  \mathbb{R}^{K_2 \times d} $, respectively.
    \begin{equation}
    \begin{aligned}
    & \bm{X} = \left\{ \bm{x}_{i} \right\}, \bm{Y} = \left\{ \bm{y}_{j} \right\}, \\
    & \left\{i \in [1,...,K_1], j \in [1,...,K_2]\right\},
    \end{aligned}
    \end{equation}
    where $d$ denotes the dimension of each node features.
    $\bm{x}_{i} \in  \mathbb{R}^{1 \times d}$ indicates the encoded features of node $ i $ in visual graph, and $\bm{y}_{j} \in  \mathbb{R}^{1 \times d}$ is the encoded features of node $ j $ in question graph.
    Next, two graph convolutional layers ($Gconv_{11}$ and $Gconv_{21}$) aggregate node features ($\bm{X}$ and $\bm{Y}$) according to the graph edge information $\bm{E}^m$ and $\bm{E}^n$, respectively.
    The parameters of $Gconv_{11}$ and $Gconv_{21}$ are shared. The $Gconv_{11}$ layer for visual graph is formulated as
    \begin{equation}
    \begin{aligned}
    \bm{X}' = \sigma(\bm{W}_{1}(\bm{X} + \bm{W}_{2}(\bm{L}^m \bm{X}))),
    \end{aligned}
    \end{equation}
    where $ \bm{W}_1 $ and $ \bm{W}_2 $ are the layer-specific trainable weight matrix, $ \sigma(\cdot) $ denotes a non-linear activation function. $ \bm{L}^m $ is a symmetric normalized laplacian of size $ K_1 \times K_1$ which could be defined as
    \begin{equation}
    \begin{aligned}
    \bm{L}^m = & (\bm{D}^{m})^{-1/2} \bm{E}^m (\bm{D}^{m})^{1/2}, \\
    \end{aligned}
    \end{equation}
    \begin{equation}
    \begin{aligned}
    \bm{D}^m = diag\left(\sum_{j=1}^{K_1} e_{0j}, \dots, \sum_{j=1}^{K_1} e_{ij}, \dots, \sum_{j=1}^{K_1} e_{K_1j}\right),
    \end{aligned}
    \end{equation}
    where $\bm{D}^m$ is a diagonal matrix, $e_{ij}$ belongs to $\bm{E}^m$. Symmetric normalized Laplacian matrix is a variant of Laplacian matrix, which can effectively aggregate information between graph nodes according to the edge matrix of graph.
    Then, the aggregated question graph features $\bm{Y}' \in \mathbb{R}^{K_2 \times d}$ can be obtained by the $Gconv_{21}$ layer in the same way. In this way, we obtain the explicit intra-modality representation $\bm{X}'$ and $\bm{Y}'$.

    Given $\bm{X}'$ and $\bm{Y}'$, the successive implicit relationship inference stage
    constructs connections between nodes by comparing their similarities. In this way, the intra-modality messages among the graph nodes can be propagated.
    Depending on this propagation, our \textit{GMA} module is capable of learning semantic relations inside the image and the question implicitly.
    In the adjacency layers (denoted as $Adj$ and shown in Fig.\ref{fig3}), the adjacency matrix $\bm{A}^m$ of visual graph and $\bm{A}^n$ of question graph are learned by calculating the similarity inside the graph node features $\bm{X}'$ and $\bm{Y}'$, respectively.
    The adjacency layer for the visual graph can be formulated as
    \begin{equation}
    \begin{aligned}
    a^m_{ij} =  \frac{\exp^{\mathcal{SIM}(\bm{x}_{i}',\ \bm{x}_{j}')}}{\sum_{j \in K_1} \exp^{\mathcal{SIM}(\bm{x}_{i}',\ \bm{x}_{j}')} },
    \end{aligned}
    \end{equation}
    \begin{equation}
    \begin{aligned}
    \mathcal{SIM}(\bm{x}_i',\bm{x}_j') = \parallel \bm{x}_i' - \bm{x}_j' \parallel_2^2,
    \end{aligned}
    \end{equation}
    where $a^m_{ij} \in \bm{A}^m$, $\bm{x}_{i}', \bm{x}_{j}' \in \bm{X}'$ and $i,j\in [1, K_1]$.
    The function $\mathcal{SIM}$ calculates the Euclidean distance between two nodes in a graph. The adjacency matrix $\bm{A}^n\in  \mathbb{R}^{K_2 \times K_2}$ can be calculated in the same manner.
    In order to improve the understanding and reasoning ability of the VQA model, another two graph convolutional layers ($Gconv_{12}$ and $Gconv_{22}$) transfer semantic relation information into the feature of each graph node according to the adjacency matrices.
    The parameters of $Gconv_{12}$ and $Gconv_{22}$ layers are also shared, and the $Gconv_{12}$ is calculated by
    \begin{equation}
        \begin{aligned}
        \bm{X}'' = \bm{X}' + \bm{W}_{3}(\bm{A}^m\bm{X}'),
        \end{aligned}
    \end{equation}
    where $\bm{W}_{3} \in  \mathbb{R}^{d \times d}$ is the layer-specific trainable weight matrix, $ \bm{X}'' \in \mathbb{R}^{K_1 \times d}$ is the visual knowledge which is outputted from the graph encoder. In a similar way, the textual knowledge $ \bm{Y}'' \in  \mathbb{R}^{K_2 \times d}$ is calculated as
    \begin{equation}
        \begin{aligned}
        \bm{Y}'' = \bm{Y}' + \bm{W}_{3}(\bm{A}^m\bm{X}'),
        \end{aligned}
    \end{equation}
    By the dual-stage graph encoder, the two modality graphs can be termed as
    \begin{equation}
        \begin{aligned}
        & \mathcal{G}_1^{'} =\{\bm{X}'', \bm{E}^{m}, \bm{A}^{m}\}, \\
        & \mathcal{G}_2^{'} =\{\bm{Y}'', \bm{E}^{n}, \bm{A}^{n}\}.
        \end{aligned}
    \end{equation}

\subsubsection{Bilateral Matching Attention}
    After the dual-stage graph encoder module, the bilateral matching attention module is proposed for cross-modal information interaction between two graphs. The nodes of the visual graph represent the visual features of all objects detected in the image, while the nodes of the question graph are the related word features of some specific objects in the question sentence. The two different modal features need to be aligned first, and then further fused, so as to achieve more accurate reasoning in the process of cross-model interaction. Inspired by the idea of graph matching algorithm, the proposed bilateral matching attention module first performs the matching operation between two graphs, and then the designed attention algorithm according to the matching scores to complete the cross-modal feature fusion and relational reasoning.
    The proposed matching attention sub-module consists of a graph matching layer and a bilateral cross-modality attention layer. 
    In the graph matching layer, an affinity matching function $f_a$ learns the graph matching score $ s_{ij} $ between the two nodes' features $ \bm{x}_{i}''\in \bm{X}'' $ and $ \bm{y}_{j}'' \in \bm{Y}'' $. The formula of the matching function $f_a$ is defined as
    \begin{equation}
        \begin{aligned}
        s_{ij} = f_a(\bm{x}_{i}'', \bm{y}_{j}''), \quad \left\{i \in [1, K_1], j \in [1, K_2] \right\}.\\
        \end{aligned}
    \end{equation}
    $f_a$ can be a bi-linear mapping followed by an exponential function which ensures all elements positive.
    \begin{equation}
        \label{equ_match_aff_S}
        \begin{aligned}
        s_{ij} & = \exp{\left( \frac{\bm{x}_{i}'' \bm{A_w}' (\bm{y}_{j}'')^T}{\tau}\right) } \\
    	   & = \exp{\left( \frac{\bm{x}_{i}'' (\bm{A_w} + \bm{A_w}^T) (\bm{y}_{j}'')^T}{2\tau}\right) },
        \end{aligned}
    \end{equation}
    \begin{equation}
        \begin{aligned}
        \left \{ \forall\ i \in [1, K_1],\ \bm{x}_{i}'' \in \mathbb{R}^{1 \times d}, \forall\ j \in [1, K_2], \ \bm{y}_{j}'' \in \mathbb{R}^{1 \times d} \right\}
        \nonumber
        \end{aligned}
    \end{equation}
    where $\tau$ is a hyper parameter, and  $ \bm{A_w} \in \mathbb{R}^{d \times d}$ denotes the learnable weights of the affinity function.
    From Eq.~\ref{equ_match_aff_S}, the graph matching affinity matrix $  \bm{S} = \left\{s_{ij}\right\} \in  \mathbb{R}^{K_1 \times K_2}$ retains the matching score between two nodes coming from two different modalities.

    In the bilateral cross-modality attention sub-module, we  merge the cross-modality features in two ways (form $\bm{X}''$ to $\bm{Y}''$ and from $ \bm{Y}''$ to $\bm{X}''$) for interactive learning by applying a softmax function.
    It outputs two cross-modality attention maps, $\bm{S}_1 \in \mathbb{R}^{K_1 \times K_2}$ and $\bm{S}_2 \in \mathbb{R}^{K_2 \times K_1}$, which separately represent the  matching degree between the two cross-modality graph nodes.
    Therefore, the node features of the visual and textual modalities can be updated by the attention maps $\bm{S}_1$ and $\bm{S}_2$:
    \begin{equation}
        \begin{aligned}
        \bm{V}^n = \bm{W}_4(\bm{Y}'' \oplus \bm{S}_1 \bm{X}'')
        \end{aligned}
    \end{equation}
    \begin{equation}
        \begin{aligned}
        \bm{V}^m = \bm{W}_5(\bm{X}'' \oplus \bm{S}_2 \bm{Y}'')
        \end{aligned}
    \end{equation}
    where $\bm{V}^m \in  \mathbb{R}^{K_1 \times d}$ and $\bm{V}^n \in  \mathbb{R}^{K_2 \times d}$ are two updated graph nodes, the operator $\oplus$ represents matrix concatenation. $\bm{W}_4 \in  \mathbb{R}^{d \times d}$ and $\bm{W}_5 \in  \mathbb{R}^{d \times d}$ are the layer-specific trainable weight matrices.

\begin{algorithm}[htb]
  \caption{Framework of our Graph Matching Attention.}
  \label{alg:Framwork}
  \begin{algorithmic}[1]
    \Require
      The input images $I$ and questions $Q$;
    \Ensure
      classifier results $scores$;
    \State Extracting the features $\bm{R}$ and $\bm{B}$ with the help of Faster RCNN algorithm; Extracting the features $q$ with the help of Bi-GRU algorithm; Extracting the syntax dependency tree $T$ through the dependency parsing algorithm;
    \label{code:fram:extract}
    \State Building the visual graph $\mathcal{G}_{1}= \{\bm{V}^m, \bm{E}^m\}$ and question graph $\mathcal{G}_{2} = \{\bm{V}^n, \bm{E}^n\}$ through the graph construction module;
    \label{code:fram:build}
    \For{stack $i \to N$}
            \State $X, Y = FC(\bm{V}^m, \bm{V}^n)$
            \State $X^{'}, Y^{'} = Gconv_{1*}\{(X, \bm{E}^m); (Y, \bm{E}^n)\}$
            \State Calculate the the adjacency  matrix $A^{m}$ and $A^{n}$ through the function $\mathcal{SIM}$;
            \State $X^{''}, Y^{''} = Gconv_{2*}\{(X^{'}, \bm{A}^m); (Y^{'}, \bm{A}^n)\}$
            \State Calculate the graph matching matrix $S$ through the affinity matching function $f_a$ and get two cross-modality attention maps $S_1$ and $S_2$;
            \State $ \bm{V^{n}}, \bm{V^{m}} = update(X^{''}, Y^{''},S_1, S_2)$
    \EndFor
    \State Get the reasoning feature $h$ through the global pooling;
    \State Calculate the Classifier results $scores = MLP(q,h)$;
    \Return $scores$;
  \end{algorithmic}
\end{algorithm}

\subsection{Answer Prediction Layer}
\label{Sec: answer prediction layer}
    After one or several stacked \textit{GMA} modules, an answer prediction module concatenates the visual and textual modal information to predict the final answer.
    Graph node features ($\bm{V^m}$ and $\bm{V^n}$) are integrated and generate a new representation $\bm{H} \in \mathbb{R}^{K_3 \times d}$ by
    \begin{equation}
        \begin{aligned}
        \bm{H} = \delta{(\bm{V}^m \oplus \bm{V}^n)},
        \end{aligned}
    \end{equation}
    where $ \delta $ denotes a nonlinear activation function, $ \oplus $ denotes the concatenating operation, and $K_3 = K_1 + K_2 $.
    Then, $ \bm{H} $ is operated by a maximum pooling and outputs the reasoning feature $\bm{h} \in \mathbb{R}^{1 \times d} $ along the node dimension.
    During question graph construction, we use Bi-GRU to embed question $Q$ into question feature $ \bm{q} \in \mathbb{R}^{1 \times d}$.
    $\bm{q} $ and $\bm{h}$ are merged by an element-wise production, and sent into a two-layer MLP to predict the scores of candidate answers. Formally,
    \begin{equation}
        \begin{aligned}
        scores = \sigma{(\mathbf{MLP}( \bm{q} \circ \bm{h}))}.
        \end{aligned}
    \end{equation}
    $N_a$ denotes the number of all possible answers, the MLP has 2048 and $N_a$ units in its first and second layer.
    In addition, we use the multi-label soft loss function to compute the soft target score of each class:
    \begin{equation}
    \begin{aligned}
    L(t, & y) = \\
    	 & \sum_{i} \left\{t_i \log\left(\frac{1}{1 + e^{-y_i}}\right) + (1 - t_i) \log\left(\frac{e^{-y_i}}{1 + e^{-y_i}} \right) \right\}
    \end{aligned}
    \end{equation}
    where $y$ is the output of our model. $t = \frac{number \ of \ votes}{nc}$ denotes the target score of each class, $nc$ is the total answer category of the corresponding dataset, which can be found in experimental implementation part in Section~\ref{section_experimental}. The detailed process of our GMA framework is expressed in algorithm~\ref{alg:Framwork}.

\section{Experiments}
    In this section, we conduct ablation studies to verify the effectiveness of each module in our proposed Graph Match Attention network, and make a comparison with state-of-the-art methods on the VQA 2.0~\cite{2018Making} and the GQA datasets~\cite{Hudson_2019_CVPR}.

\subsection{Dataset and Metrics}
    \subsubsection{The VQA 2.0 dataset}
    VQA 2.0 dataset~\cite{2018Making} is built on top of the widely-used VQA 1.0 dataset~\cite{Antol_2015_ICCV} and it contains a total of 1,105,904 questions and 204,721 images collected from the COCO dataset~\cite{2015Microsoft}. Each question has 10 answers labeled by 10 AMT workers. The sizes of \textit{train} split, the \textit{val} split and the \textit{test} split are 443,757, 214,354 and 447,739, respectively. The dataset has two types of questions, \emph{i.e.} the open-ended and the multiple-choice questions. We focus on the open-ended task for common comparisons.
    After preprocessing the dataset, we take the answers that appear more than 9 times in the \textit{train} split as candidate answers and generate 3129 candidate answers as the final classification category. We train our model on the \textit{train} split and evaluate on the \textit{val} split. When testing on \textit{test-dev}, \textit{train}, \textit{val} splits and extra Visual Genome dataset~\cite{Ranjay_2017_IJCV} are combined for training.
    VQA 2.0 datasets report accuracy on three categories of questions: \textit{yes/no}, \textit{number} and \textit{other}.
    The ground truth score of an answer is determined by voting among 10 human annotators:
    \begin{equation}
        \begin{aligned}
        ACC(ans) = min(1, \frac{\#humans\ provided\ ans}{3})
        \end{aligned}
    \end{equation}
    \noindent
    \subsubsection{The GQA dataset}
    GQA dataset~\cite{Hudson_2019_CVPR} is a new dataset for the Visual Question Answering task over real-world images.
    Compared with VQA 2.0 dataset, GQA needs more reasoning skills, such as spatial understanding, multi-step reasoning, and so on. In the GQA dataset, 52\% of the questions need the trained model to understand the visual relationship and conduct relational reasoning.
    The effectiveness of our model can be verified directly by reasoning on these complex questions.
    Apart from the standard \textit{Accuracy} metric measured on several different categories of questions, the dataset also provides some other metrics to better evaluate the performance of the model. The \textit{Consistency} metric is used to measure the level of consistency in responses across different questions. The \textit{Validity} metric measures whether the model gives valid answers, ones that can be theoretically correct for the question. The \textit{Plausibility} metric measures whether the model responses are reasonable in the real world or not making sense. The \textit{Distribution} metric measures the difference between the predicted answer distribution and the ground truth.

\subsection{Experimental Implementation}
\label{section_experimental}
\begin{table}[htp]
    \centering
    \small
    \setlength{\tabcolsep}{1mm}{
    \caption{Ablation studies of our proposed Graph Match Attention network on the VQA 2.0 \textit{val} split.} \vspace{0.2cm}
    \label{tab:1}
    \begin{spacing}{1.3}{
    \begin{tabular}{l c c}
        \hline
        Component & Setting & Accuracy \\
        \hline
        BUTD~\cite{Anderson_2018_CVPR}
        & Bottom-up & 63.37 \\
        \hline
        \multirow{3}{3cm}
        {BAN~\cite{NIPS2018_7429}}
            & {BAN-1} & 65.36 \\
            & {BAN-4} & 65.81 \\
            & {BAN-12} & 66.04 \\
        \hline
        \multirow{3}{3cm}{DFAF~\cite{Gao_2019_CVPR}}
            & {DFAF-1} & 66.21 \\
            & {DFAF-5} & 66.58 \\
            & {DFAF-8} & 66.60 \\ \hline
        Default & {GMA-1} & \textbf{66.80} \\
        \hline
        \multirow{2}{3cm}{\# of stacked blocks}
            & {GMA-2} & 67.34 \\
            & {GMA-3} & \textbf{67.47} \\
            & {GMA-4} & 67.41 \\
        \hline
        Part of attention & Bilateral Matching Attention & 66.36 \\
        \hline
        \multirow{3}{3cm}{Part of \\ graph encoder}
            & only Explicit Stage & 66.67\\
            & only Implict Stage & 66.64\\
            & Dual-Stage & 66.80 \\
        \hline
        \multirow{2}{3cm}{Cross-graph \\ feature fusion}
            & Addition & 66.56 \\
            & Concatenation & 66.80\\
        \hline
        \multirow{3}{3cm}{Bounding Box \\ Embedding}
            & None & 66.35\\
            & Absolute Position & 66.67 \\
            & Normalized Position &  66.80\\
        \hline
    \end{tabular}}
    \end{spacing}}
\end{table}
\begin{table}[htp]
    \centering
    \small
    \setlength{\tabcolsep}{0.8mm}{
    \caption{Ablation studies of stacked GMA modules on the GQA \textit{val} split.} \vspace{0.2cm}
    \label{tab:2}
    \begin{spacing}{1.3}{
    \begin{tabular}{l c c c c c}
        \hline
        Component & SceneGCN~\cite{YangScene}  & LCGN~\cite{DBLP:journals/corr/abs-1905-04405} & GMA-1 & GMA-2 & GMA-3  \\
        \hline
        Accuracy & 62.45 & 63.90 & 63.93 & 64.22 & \textbf{64.55} \\
        \hline
    \end{tabular}}
    \end{spacing}}
    \vspace{-2mm}
\end{table}
\subsubsection{Feature preprocessing}
For visual feature extraction, we adopt Faster RCNN~\cite{NIPS2015_5638} pre-trained in Visual Genome~\cite{Ranjay_2017_IJCV} to detect object proposals, extracting their bounding boxes $\bm{B} \in \mathbb{R}^{K_1 \times 4}$ and the RoI features $\bm{R} \in \mathbb{R}^{K_1 \times 2048}$. Those features contain 10 to 100 object proposals per image and we pad the number of object proposals to the fixed size $K_1=100$. In the visual graph construction module, in order to make the graph node features contain spatial information,
$\bm{B}$ is normalized by the image size and concatenated with $\bm{R}$ to form the visual graph nodes $\bm{V}^m$.
In question processing, we set the sentence length to $K_2=14$ and embed it as a sequence of 300-dimensional word vectors by a pre-trained GloVe embedding function. Next, this embedded feature is sent into a Bi-GRU~\cite{DBLP:journals/corr/ChungGCB14} layer to produce a global feature $\bm{q} \in \mathbb{R}^{1\times2048}$.
Simultaneously, we adopt the dependency parsing algorithm~\cite{manning-etal-2014-stanford} to analyze the syntax dependency tree in the given question. Based on the constructed tree, we then put all the related words of each word in the corresponding question together as a node of the question graph. Therefore, each graph node has a variable number of words. Next, we map those words into a sequence of GloVe word embedding features and also use the Bi-GRU module to extract the features $\bm{V}^n \in  \mathbb{R}^{K_2\times2048}$.

\subsubsection{Hyper-parameters}
In the visual graph construction module, we set the IoU threshold to 0.3 to link sparse image regions. 
In the \textit{GMA} module, all the hidden dimensions $d$ are set to 2048. The hyper-parameter $\tau$ is set as $1/\sqrt{2028}$. The final answer category $nc$ of our \textit{GMA} model is set to 3129 classes on VQA 2.0 dataset and 1834 classes on \textit{GQA} dataset. During training, the dropout layer is applied to the pre-trained GloVe word embedding features and question features with a 0.25 probability, and is applied to image features and final reasoning features with a 0.5 probability.
The model is trained by Adamax optimizer for 35 epochs with a batch size of 256.
The learning rate is initialized as $5e^{\\-4}$ and linearly adjusted by $2e^{\\-3}$ at epoch 4.
After 25 epochs, the learning rate is half decreased every 2 epochs.

\subsection{Ablation study of GMA}
    In this section, we conduct ablation studies on VQA 2.0 \textit{val} split to verify the validity of each part of our Graph Matching Attention network. We also compare the effects of different hyper parameters on \textit{GMA} model by visualization. In addition, we conduct ablation studies on GQA \textit{val} split to verify the effectiveness of stacked \textit{GMA} modules.
    \begin{figure}
    \begin{center}
    \centering 
    \includegraphics[scale=0.55]{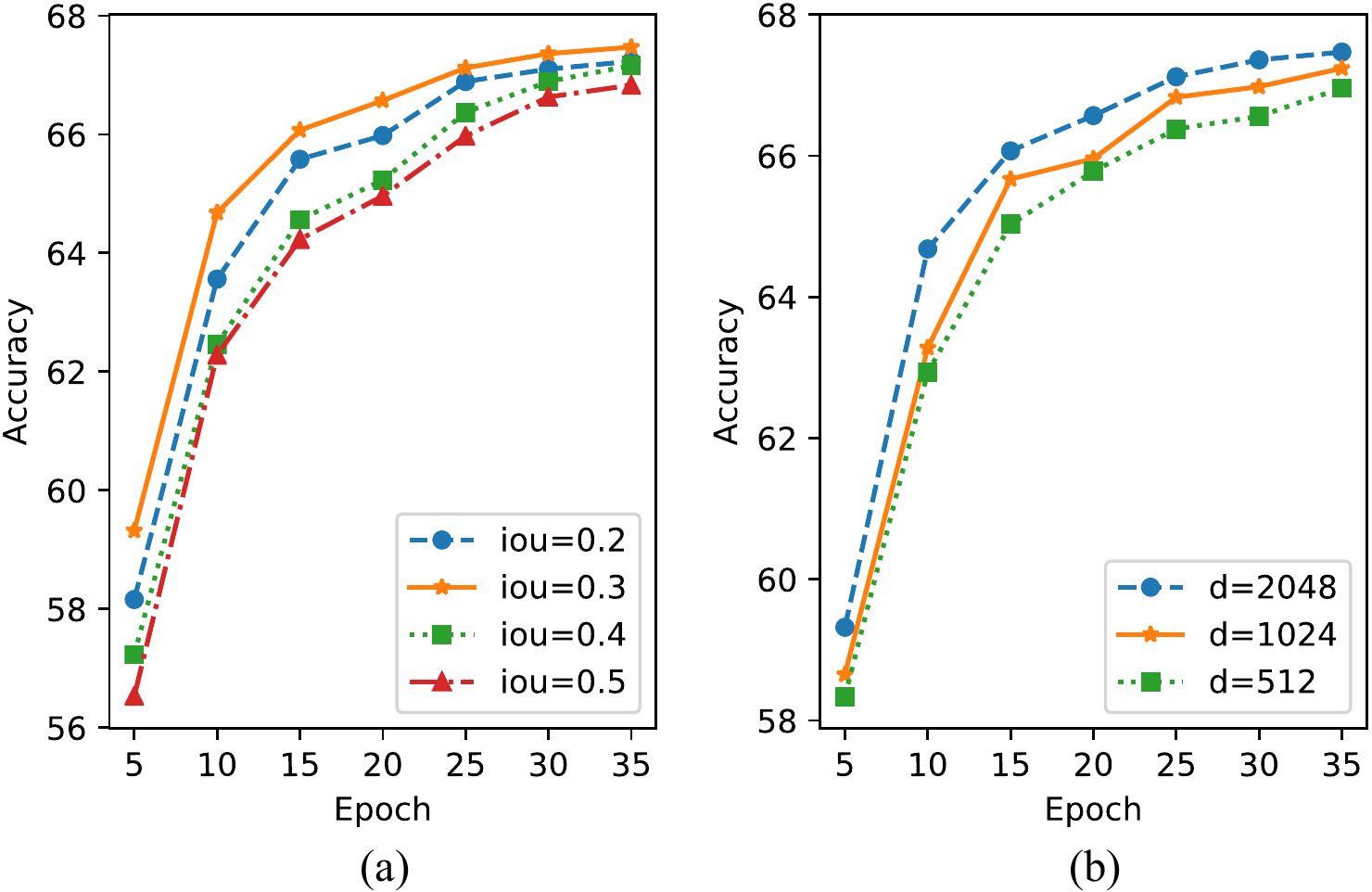}
    \end{center}
    \caption{The visualization results of \textit{GMA} model training process reflect the influence of different hyper-parameter settings on the model performance. These results were trained on VQA 2.0 train split and evaluated on  the \textit{val} split. Different curves in the left (a) represent different IoU thresholds, and the right (b) represent different hidden dimensions $d$ of the model.
    }
    \label{fig6}
    \vspace{-3mm}
    \end{figure}
    \subsubsection{Comparison to Attention-based model}
    The goal of those experiments is to compare our approach with strong baselines for real VQA in controlled conditions. As shown in Table.~\ref{tab:1},
    the default setting of our network contains only one \textit{GMA} module (\textit{GMA-1}). And we compare our GMA network with several famous attention-based methods, \emph{i.e.} the BUTD~\cite{Anderson_2018_CVPR}, the BAN~\cite{NIPS2018_7429} and the DFAF~\cite{Gao_2019_CVPR}.
    For fair comparisons, we use the same bottom-up features to insure competitiveness.
    In terms of the attention mechanism comparison, our \textit{GMA-3} outperforms the BAN-12 by 1.43\% and DFAF-8 by 0.87\% on the \textit{Accuracy} metric.
    Our \textit{GMA} model reaches a higher accuracy than those three attention-based methods on VQA 2.0 \textit{val} split. It demonstrates our graph matching attention network answers questions correctly and generates more effective relational reasoning.

\subsubsection{Number of stacked \textit{GMA} modules}
    In order to carry out deeper relationship reasoning, we stack several \textit{GMA} modules. In this section, we investigate the influence of the number of the stacked \textit{GMA} modules on VQA 2.0 dataset.
    As shown in the 5th setting ("\# of stacked blocks") of Table~\ref{tab:1}, compared with \textit{GMA-1}, stacking multiple \textit{GMA} modules improves the performance.
    For example, \textit{GMA-2} improves the results by 0.54\%, and \textit{GMA-3} gains a further improvement of 0.67\%. With the growing number of stacked \textit{GMA} modules, \textit{GMA-4} turns up with a slightly performance degradation of 0.06\%, compared with \textit{GMA-3}. It can be interpreted as the emergency of over-smoothing when message passing in deep graph convolutional networks.
    We also verify the effectiveness of the stacking operation on the GQA dataset, as shown in Table \ref{tab:2}. The \textit{GMA-3} overperform the \textit{GMA-1} by 0.62\%. According to the experimental results, with the increasing number of \textit{GMA} modules, the ability of the network  to answer questions that require reasoning is enhanced. By cascading several \textit{GMA} modules, the network achieves better relationship reasoning.
\begin{table}[htp]
    \centering
    \small
    \caption{Comparison of our \textit{GMA} with state-of-the-art on  \textit{test-dev} and  \textit{test-std} of the VQA 2.0 dataset.}
    \label{tab:3}\vspace{0.2cm}
    \begin{spacing}{1.3}{
    \begin{tabular}{l c c c c c }
    	\hline
        \multirow{2}{*}{Model} & \multicolumn{4}{c}{test-dev}  & test-std  \\
        & all & Y/N & Num & Other & all \\ \hline
        \multicolumn{2}{l}{Attention Learning} \\
        \hline
        SAN~\cite{Yang2016Stacked} & 58.7 &  79.3 & 36.6 &  46.1 &  58.9 \\
        FDA~\cite{DBLP:journals/corr/IlievskiYF16} & 59.2 & 81.1 & 36.2 & 45.8 & 59.2 \\
        Co-attention~\cite{LuHierarchical} & 61.8  & 79.7 & 38.7 & 51.7 & 62.1 \\
        BUTD~\cite{Anderson_2018_CVPR} & 65.32 & 81.82 & 44.21 & 56.05 & 65.67 \\
        RelAtt~\cite{DBLP:journals/corr/abs-1805-09701} & 65.69 &  83.55 & 36.92 & 56.94 & - \\
        DCN~\cite{DBLP:journals/corr/abs-1804-00775} & 66.89 & 84.61 &  42.35 & 57.31 & 67.02 \\
        RAF~\cite{DBLP:journals/corr/abs-1805-04247} & 67.2 & 84.1 & 44.9 &  57.8 & 67.4 \\
        BAN+Glove~\cite{NIPS2018_7429} & 69.52 & 85.46 & 50.66 & 60.50 & - \\
        \hline
        \multicolumn{2}{l}{Graph Learning} \\ \hline
        GraphLearner~\cite{DBLP:journals/corr/abs-1806-07243} & 65.89 & 82.91 & 46.13 & 55.22 & 66.18 \\
        SceneGCN~\cite{YangScene} & 66.81 & 82.72 & 46.85 & 57.77 & 67.14 \\
        VCTREE-HL~\cite{DBLP:journals/corr/abs-1812-01880}  & 68.19 & 84.28 & 47.78 & 59.11 & 68.49 \\
        \hline
        \multicolumn{2}{l}{Relation Learning} \\ \hline
        MUTAN~\cite{Ben-younes_2017_ICCV} & 66.01 & 82.88 & 44.54 & 56.50 & - \\
        Pythia~\cite{DBLP:journals/corr/abs-1807-09956} & 68.05 & - & - & - & - \\
        MFH~\cite{Yu2017Beyond} & 68.76 & 84.27 & 50.66 & 60.50 & - \\
        Free\_VQA~\cite{Yiyi2019Free} & 67.53 & 83.96 & 46.50 & 58.29 & 67.65 \\
        BLOCK~\cite{DBLP:journals/corr/abs-1902-00038} & 67.58 & 83.6 & 47.33 & 58.51 & 67.92 \\
        SEADA~\cite{2020Semantic} & 67.78 & 84.08 &  47.55 & 58.48 & 68.27 \\
        MuRel~\cite{Cadene_2019_CVPR} & 68.03 & 84.77 & 49.84 & 57.85 & 68.41\\
        Counter~\cite{DBLP:journals/corr/abs-1802-05766} & 68.09 & 83.14 & 51.62 & 58.97 & 68.41 \\
        DF~\cite{Chenfei2019Differential} & 68.31 & 84.33 & 48.2 & 59.22 & 68.59 \\
        \hline
        Ours(\textit{GMA}) & \textbf{69.86} & \textbf{85.62} & \textbf{51.74} & \textbf{60.51} & \textbf{70.16} \\
        \hline
    \end{tabular}}
    \end{spacing}
\end{table}
\begin{figure*}
    \begin{center}
    \centering
    \includegraphics[width=0.9\textwidth]{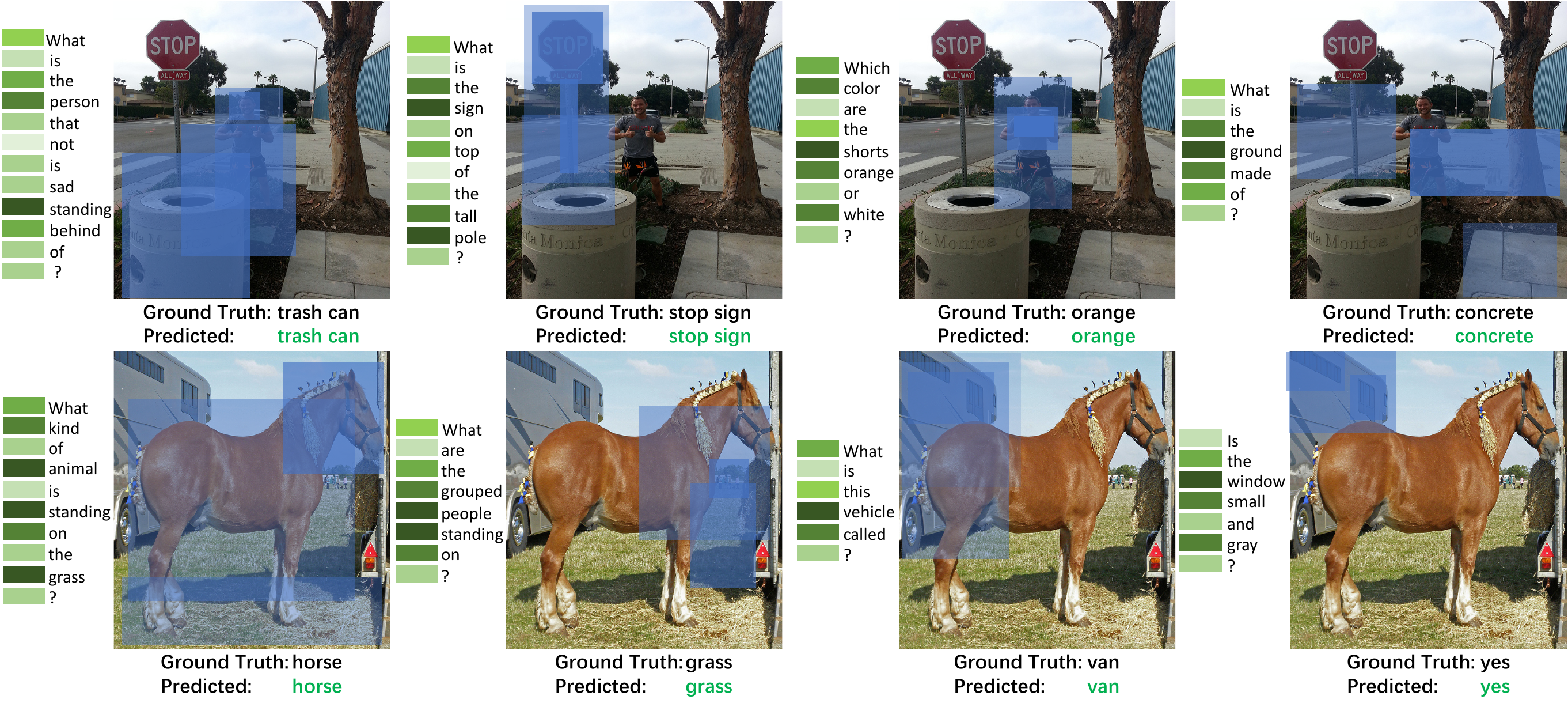}
    \end{center}
    \caption{Qualitative results of our \textit{GMA} network on GQA dataset. Each line visualizes different questions corresponding to the same image. The left side of each image represents the question, the bottom side represents ground truth and predicted answer.
    }
    \label{fig5}
\end{figure*}
\begin{table*}[htp]
    \caption{Comparison of our method with state-of-the-art methods on the GQA test split.} \vspace{0.2cm}
    \centering
    \begin{spacing}{1.1}{
    \resizebox{\textwidth}{!}{
    \begin{tabular}{ l | c | c | c | c | c | c | c |c | c}
    	\hline
         & Global Prior
         & Local Prior
         & CNN+LSTM
         & BUTD
         & MAC
         & SceneGCN
         & LCGN
         & GRNs
         & Ours(\textit{GMA}) \\
        \hline
        Open & 16.52 & 16.99 & 31.8 & 34.83 & 38.91 & 40.63 & - & 41.24 &\textbf{42.3} \\
        Binary & 42.99 & 47.53 & 63.26 & 66.64 & 71.23 & 70.33 & - & \textbf{74.93} & 73.2\\
        Distribution & 130.86 & 21.56 & 7.46 & 5.98 & 5.34 & 6.43 & - & 5.76 & \textbf{5.56} \\
        Validity & 89.02 & 84.44 & 96.02 & 96.18 & 96.16 & 95.9 & - & 96.14 & \textbf{96.41} \\
        Plausibility & 75.34 & 84.42 & 84.25 & 84.57 & 84.48 & 84.23 & - & 84.68 & \textbf{85.05} \\
        Consistency & 51.78 & 54.34 & 74.57 & 78.71 & 81.59 &  83.49 & - & \textbf{87.41} & 83.95 \\
        \hline
        Accuracy & 28.93 & 31.31 & 46.55 & 49.74 & 54.06 & 54.56 & 56.10 & 57.04 & \textbf{57.26} \\
        \hline
    \end{tabular}
	}}
	\end{spacing}
    \label{tab:4}
\end{table*}

\subsubsection{Components of \textit{GMA} module}
In order to validate the effectiveness of each part of our \textit{GMA} module, some modules are deleted from \textit{GMA-1} model. We first delete the Dual-Stage Graph Encoder sub-module and  
the result is reported in the 6th setting of Table~\ref{tab:1}.
Compared with the default setting, the performance without the Dual-Stage Graph Encoder decreases by 0.44\%, reflecting the significant role of intra-modality reasoning. Because the Bilateral Matching Attention sub-module can intuitively align features between the visual and language modalities, the attention could learn better relational semantics depending on the graph matching operation. Even though, the performance of the \textit{GMA} without the graph encoder outperforms the BUTD~\cite{Anderson_2018_CVPR} and BAN-1~\cite{NIPS2018_7429} by 2.99\% and 1.0\%, respectively.
The large gaps validate the advantage of our graph match attention mechanism. Then, we design fine granularity studies on the part of graph encoder. As shown in the 7th component of Table~\ref{tab:1}, the result of the Explicit Stage Graph Encoder was improved by 0.31\% due to that the explicit structure information in the two modalities can be integrated into each graph node.
Besides, merely using the Implicit Stage Graph Encoder, the performance improves 0.28\%.
Since the implicit relationships between each graph node have been learned, the performance can be promoted by the matching attention sub-module.
Moreover, the Dual-stage Graph Encoder can be enhanced by 0.43\%, indicating the complementarity between the explicit and the implicit graph structure information.
\subsubsection{strategies and Hyper-parameters}
The strategies of the cross-graph feature fusion and the embedding methods of the bounding box are also be investigated. As illustrated in the 8th component of Table \ref{tab:1}, we test two different fusion strategies, concatenation and addition. The concatenation strategy combines two different modalities features through concatenate operation, and then updates features through the full connection layers. And the addition strategy simply fuse the features of two graphs by addition operation and also apply the full connection layers. The final results show that the concatenation strategy for cross-graph feature fusion is better than the addition strategy.
Besides, we also test several embedding methods of the bounding box to improve the performance of our \textit{GMA} model. In the 9th component of Table \ref{tab:1}, for the bounding box embedding, the normalized position embedding can get the best performance compared with no bounding box embedding and absolute position embedding.
At last, we investigate the hyper-parameters settings (as demonstrated in Figure~\ref{fig6}). The model is trained on the VQA 2.0 \textit{train} split for 35 epochs with batch size of 256, and evaluated on the \textit{val} split. The left (a) shows the performance of IoU threshold used in the graph construction module (~\ref{Sec: Graph construction}) and the right (b) is the result of the dimension $d$ in all hidden layers. It can be found that the network performs best when IoU threshold is set to 0.3 and dimension $d$ is set to 2048.

\begin{figure*}
    \begin{center}
    \centering
    \includegraphics[width=0.98\textwidth]{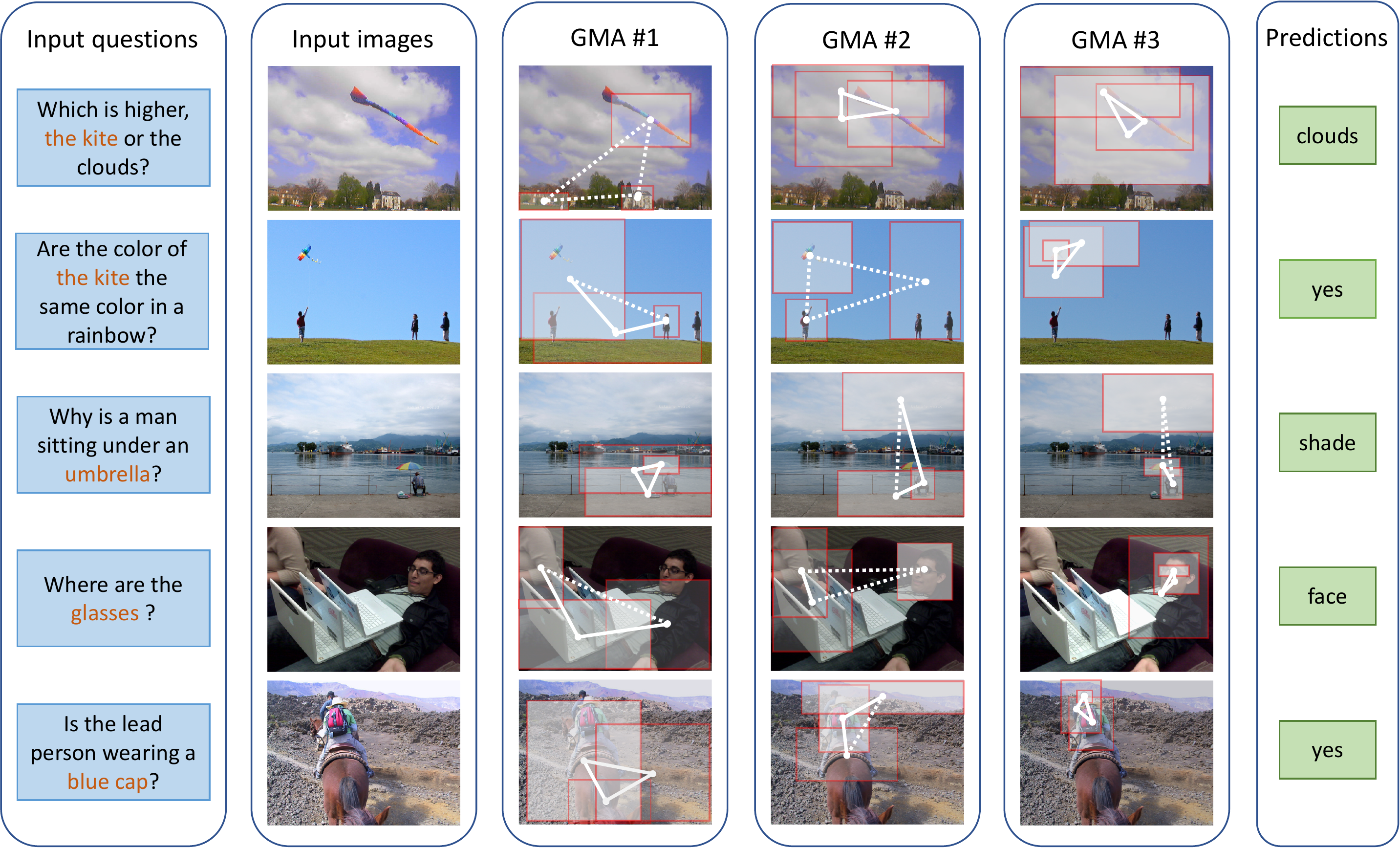}
    \end{center}
    \caption{Qualitative results of our \textit{GMA} network on VQA 2.0 dataset. \textit{GMA} $\boldsymbol{\#1}$, \textit{GMA} $\boldsymbol{\#2}$ and \textit{GMA} $\boldsymbol{\#3}$ denote the intermediate results of the \textit{GMA-3} model and Predictions are the final answers to the output of the whole model.
    }
    \label{fig4}
    \vspace{-2mm}
\end{figure*}

\subsection{Comparison with State-of-the-arts methods}

\subsubsection{Performance on the VQA 2.0 dataset}
Table \ref{tab:3} shows the comparison results of our proposed network and the state-of-the -arts on the VQA 2.0 test split.
The results demonstrates that our \textit{GMA} mechanism improves the performance over the baseline BUTD~\cite{Anderson_2018_CVPR} by 4.49\% in term of the test-std accuracy.
We also compare our model with some excellent attention-based approaches, such as SAN~\cite{Yang2016Stacked}, FDA~\cite{DBLP:journals/corr/IlievskiYF16}, Co-attention~\cite{LuHierarchical}, RelAtt~\cite{DBLP:journals/corr/abs-1805-09701}, DCN~\cite{DBLP:journals/corr/abs-1804-00775}, RAF~\cite{DBLP:journals/corr/abs-1805-04247},
and BAN+Glove~\cite{NIPS2018_7429}.
In table \ref{tab:3}, our \textit{GMA} outperforms these methods in all the metrics, which proves that the proposed Graph Matching Attention mechanism is more effective in the VQA task.
Since our model builds graph structure for both the image and the question representation, we compare our model with the graph-based and tree-based approaches, such as GraphLearner~\cite{DBLP:journals/corr/abs-1806-07243}, SceneGCN~\cite{YangScene}, VCTREE-HL~\cite{DBLP:journals/corr/abs-1812-01880}, which merely built graph on visual modality.
Our \textit{GMA} method overwhelms these methods, which verifies the cross-modality graph-based representation is feasible.
Finally, we make comparison with  state-of-the-art relationship learning methods, including Free\_VQA~\cite{Yiyi2019Free},
BLOCK~\cite{DBLP:journals/corr/abs-1902-00038},
SEADA~\cite{2020Semantic},
MuRel~\cite{Cadene_2019_CVPR},
Counter~\cite{DBLP:journals/corr/abs-1802-05766}, and DF~\cite{Chenfei2019Differential}. Results shows that our accuracy score on the test-dev split overwhelms these methods.

\subsubsection{Performance on the GQA dataset}
Table \ref{tab:4} shows the results of our network on the GQA test split.
We compare our \textit{GMA} model with both baselines as well as state-of-the-art methods. The baselines include Local Prior, Global Prior, and CNN+LSTM, which are introduced by GQA dataset~\cite{Hudson_2019_CVPR}. Specifically, based on the question group, the Local and Global Prior models return the most common answer for each group. The CNN+LSTM model is simple fusion approaches between the text and the image.
As we can see in Table~\ref{tab:4}, the performance of our \textit{GMA} model is much better than those baselines. In terms of \textit{Accuracy} metric, our \textit{GMA} model is 28.33\% higher than that of Local Prior model, 25.95\% higher than that of Global Prior model, and 10.71\% higher than CNN+LSTM model. The results verifies that our method is effective in solving visual question answering task on GQA dataset.
Next, we compare with several attention-based approaches, including  BUTD~\cite{Anderson_2018_CVPR} and MAC~\cite{Drew_2018_MAC}.
Our \textit{GMA} model exceed the BUTD model by 7.52\% and the MAC model by 3.20\%, which demonstrate that our graph match attention mechanism is available for cross-modality alignment and can effectively carry out feature fusion between different modalities.
Finally, we compare our \textit{GMA} model with several state-of-the-art graph-based methods, such as SceneGCN~\cite{YangScene}, LCGN~\cite{DBLP:journals/corr/abs-1905-04405}, and GRNs~\cite{2019Graph}. These methods applied graph message passing strategy and performed relational reasoning. Our model works better than SceneGCN and LCGN in all metrics, with the overall accuracy 2.7\% and 1.16\% higher than the two, respectively. Besides, the performance of our \textit{GMA} network is comparable with GRNs~\cite{2019Graph}. 
\begin{figure*}
    \begin{center}
    \centering
    \includegraphics[width=0.95\textwidth]{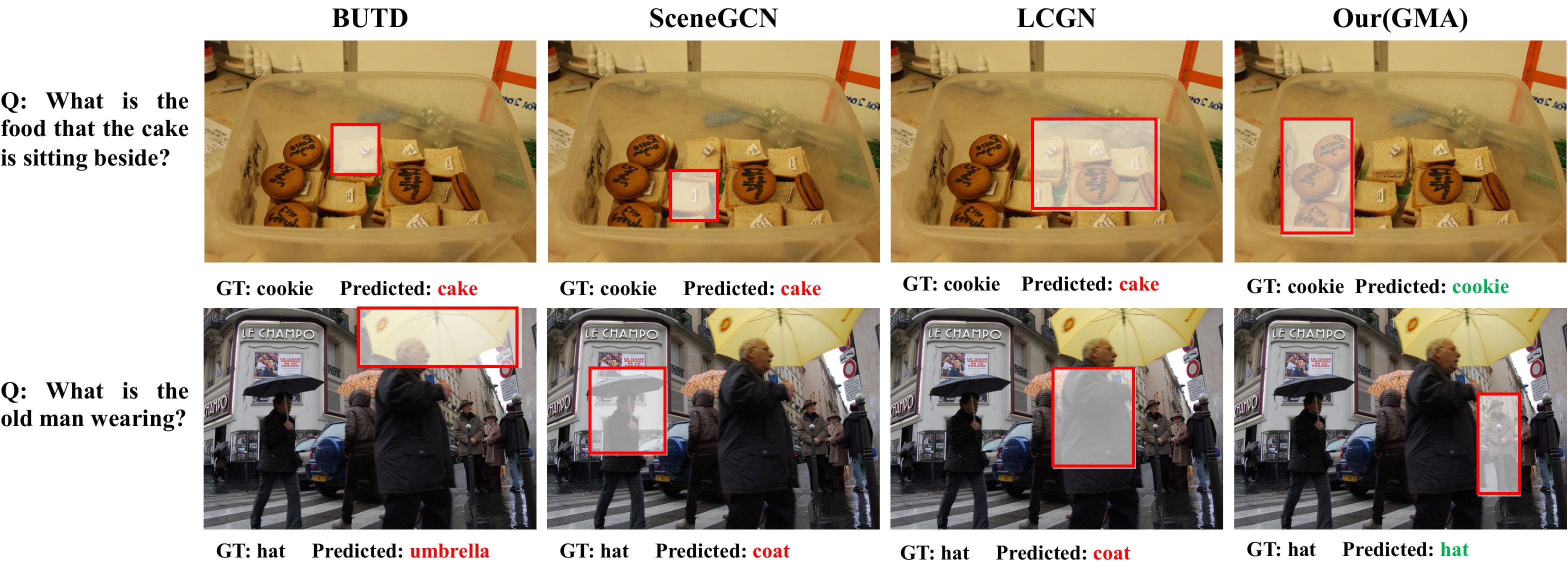}
    \end{center}
    \caption{Comparison of visualization results with state-of-the-art methods on GQA dataset. Our(\textit{GMA}) method is compared with SceneGCN and LCGN by using the same images and questions. The bottom of each image represents the prediction answers.
    }
    \label{fig7}
    \vspace{-2mm}
\end{figure*}
\subsection{Visualization}
In this section, we visualize the qualitative results of \textit{GMA} model on GQA dataset, as depict in Figure~\ref{fig5}.
Each row in the figure corresponds to the prediction results of \textit{GMA} model with different questions on the same image. The left side of each image represents the question, and the bottom side represents the ground truth and predicted answer. According to the calculated Graph Matching Attention map, we visualize each word in the question and the three most relevant objects in the image.
From Figure \ref{fig5}, the GMA can effectively understand complex questions, and find the target objects associated with the question in the image, and then can correctly answer the question through effective relational reasoning.
To better illustrate the effectiveness of our \textit{GMA} mechanism, we also visualize the learned information in each \textit{GMA} module. Figure~\ref{fig4} is the results of our trained \textit{GMA-3} model on the VQA 2.0 dataset. It finds the most relevant nodes in the question graph according to the index of the last global pooling layer (corresponding to the orange words in the input questions). We illustrate the top 3 visual objects in visual graph, which has a high correlation value (objects drawn by the red rectangles), depending on the graph match attention maps. According to the edge and affinity matrices of the visual graph, we visualize the explicit relationships (the white solid line) and the implicit relationships (the white dotted line) between the objects in a given image. It can be find that our \textit{GMA} model can find textual keywords in the questions and accurately locate visual objects in the images. As the \textit{GMA} layers are stacked, it learns much deeper relational semantics for answering questions. At last, we visualize the comparison between \textit{GMA} and state-of-the-art methods on the GQA dataset in Figure.~\ref{fig7}. The comparison results demonstrate that our \textit{GMA} network correctly solve some questions by deeper relationship reasoning.

\section{Conclusion}
In this paper, we have proposed a Graph Matching Attention (\textit{GMA}) module to solve the problem of cross-modality feature fusion and relational reasoning in VQA task by using graph matching network. Our network not only uses the whole question embedding feature, but also analyze the syntax structure of question and constructs the question graph. We apply the graph-structure representation in both the visual and textual modalities. Then, the Graph Matching Attention module learns the relationships inside the visual and textual features and propagates the cross-modality information mutually.
By stacking multiple blocks of \textit{GMA}, the inference ability will be enhanced on the VQA benchmarks.  The experimental data and the attention-based visualization provide strong evidence that our model has the capability in multi-modal feature fusion and relation reasoning.


%





\ifCLASSOPTIONcaptionsoff
  \newpage
\fi



%

\bibliographystyle{IEEEtran}

%



\vfill


\end{document}